%% file: ms.tex
\documentclass[11pt]{article} 

\usepackage{graphicx}
\usepackage{booktabs}
\usepackage{algorithm}
\usepackage{algpseudocode}
\usepackage[T1]{fontenc}
\usepackage{lmodern}
\usepackage[flushleft]{threeparttable}
\usepackage[hidelinks]{hyperref} %
\hypersetup{hidelinks}
\usepackage{multirow}
\usepackage{longtable}
\usepackage[margin=3cm]{geometry} %
\usepackage{amsmath}
\usepackage{authblk}
\usepackage{caption}
\usepackage{cite}

\title{Investigating generalization capabilities of neural networks by means of loss landscapes and Hessian analysis}
\author[1,*]{Nikita Gabdullin}
\affil[1]{Joint Stock ``Research and production company ``Kryptonite'' \authorcr
E-mail: n.gabdullin@kryptonite.ru}
\affil[*]{corresponding author}
\date{}
\begin{document}

    \captionsetup[table]{labelformat={default},labelsep=period,name={Table}}

    \maketitle

    \begin{abstract}
        This paper studies generalization capabilities of
        neural networks (NNs) using new and improved PyTorch library Loss
        Landscape Analysis (LLA). LLA facilitates visualization and analysis of
        loss landscapes along with the properties of NN Hessian. Different
        approaches to NN loss landscape plotting are discussed with particular
        focus on normalization techniques showing that conventional methods
        cannot always ensure correct visualization when batch normalization
        layers are present in NN architecture. The use of Hessian axes is shown
        to be able to mitigate this effect, and methods for choosing Hessian
        axes are proposed. In addition, spectra of Hessian eigendecomposition
        are studied and it is shown that typical spectra exist for a wide range
        of NNs. This allows to propose quantitative criteria for Hessian
        analysis that can be applied to evaluate NN's performance and assess its
        generalization capabilities. Generalization experiments are conducted
        using ImageNet-1K pre-trained models along with several models trained
        as part of this study. The experiment include training models on one
        dataset and testing on another one to maximize experiment similarity to
        model performance ``in the Wild''. It is shown that when datasets
        change, the changes in criteria correlate with the changes in accuracy,
        making the proposed criteria a computationally efficient estimate of
        generalization ability, which is especially useful for extremely large
        datasets.
    \end{abstract}

    \emph{Keywords}: Neural networks, loss landscape, Hessian analysis, vizualization, generalization. 
    
    \input{full}

\end{document}

%% file: full.tex
\section{Introduction}\label{introduction}

Neural networks (NNs) have become indispensable in many fields of
science and technology, including computer vision, robotics,
cybersecurity, medicine, and others. NNs evolved significantly from
small proof-of-concept models to large-scale industrial applications
where processing enormous amounts of new data is necessary. Modern
NNs are often trained on extremely large datasets under the assumption
that this would result in good generalization to cases ``in the Wild''~\cite{GE1,GE2}. 
However, there is still very little research into
optimal conditions that can ensure good generalization of NNs.

This increases the demand for analysis methods that can provide
additional information about NN performance other than primitive
``accuracy'' calculation. In this paper we study such method which is
associated with plotting and analyzing loss landscapes of NNs to assess
NN stability and generalization capabilities~\cite{LLO,LLe}. It allows
to explore loss variation for specific models under different conditions
which has been shown to correlate with training quality and
generalization capability. The results can be visualized in a
variety of ways and a research question regarding the optimal method is
still open.

In this paper we present a new PyTorch~\cite{Pt} library called Loss
Landscape Analysis (LLA)~\cite{LLA} which combines various approached to
loss landscape plotting with analysis methods that can provide
qualitative and quantitative assessment metrics of NN performance. We
significantly extend an older library~\cite{LLg} by adding new analysis
modes, different evaluation axes support, new weight update methods,
etc. LLA also incorporates modern techniques for Hessian analysis~\cite{HP,HG}.

Whereas the loss landscape research field is not new, there are still
very few criteria that can provide quantitative analysis metrics. To
bridge this gap, we investigate the behavior of loss landscapes and
Hessians of different NNs and formulate criteria for their analysis. We
then conduct extensive generalization experiments to assess the
viability of the proposed criteria. We also show that conventional loss
landscape analysis techniques may yield incorrect results when applied
to modern networks, showing that additional research in this field is
needed. We primarily focus on NNs trained to solve classification tasks
using supervised learning~\cite{Sr}, and all results reported in this
paper are obtained using LLA.

The rest of the paper is organized as follows: Section~\ref{methodology} discusses loss
landscape analysis methodology and highlights existing problems, Section~\ref{hes-section}
studies Hessians of conventional NNs in different regimes and proposes
criteria for Hessian analysis, Section~\ref{gen-exp} summarizes generalization
experiment results and investigates the applicability of the proposed
analysis criteria, and Section~\ref{conclusions} concludes the paper.

\section{Loss landscape analysis methodology}\label{methodology}

\begin{figure} 
    \centering
    \includegraphics[scale=0.5]{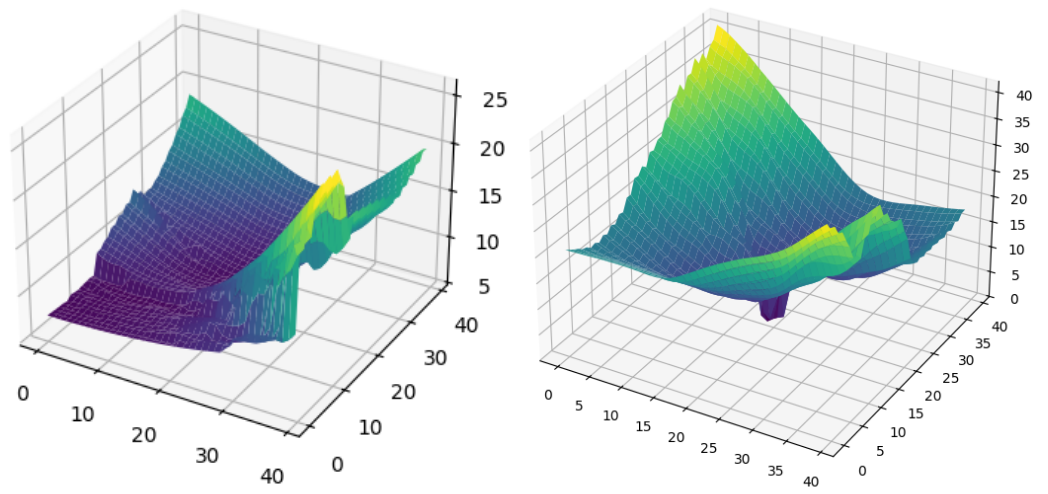}
    \caption{Loss landscapes of (left) ResNet18, and (right) VIT-small plotted along random axes.}\label{fig:2-1}
\end{figure}

Loss landscapes are obtained by varying model weights and calculating
corresponding values of a specific loss function. Since weights'
dimension is extremely large, several (typically one or two) direction
vectors are chosen to plot the landscapes, with vectors' dimensions
matching the dimensions of weights. These vectors commonly have all
values randomly taken from uniform or normal distribution~\cite{LLO,LLg},
and they are referred to as random directions. Figure~\ref{fig:2-1} shows
loss landscapes for convolutional NN (CNN) ResNet18~\cite{RN} and Visual
Transformer (VIT)~\cite{VT} plotted along random directions.

However, one should be mindful of the scale at which loss landscapes are
plotted, since this can impact the way its surface is perceived. The
most common weight update equation used for loss calculation is

\begin{equation}
    L = f\left(w + a \cdot d_{\text{1}} + b \cdot d_{\text{2}} \right),
    \label{eq:1}
\end{equation}

where \emph{w} are the original weights, and \emph{a} and \emph{b} are
coefficients corresponding to direction vectors
\emph{d\textsubscript{1}} and \emph{d\textsubscript{2}}, respectively.
Loss landscape is plotted by varying \emph{a} and \emph{b} in steps and
recording the resulting loss value. It should be noted that the choice
of the range for \emph{a} and \emph{b}, which defines how much the
weights would change, is somewhat arbitrary. In LLA we follow the
procedure proposed in~\cite{LLg} that uses 40 steps in [-20,20] range
for both coefficients, with -20 corresponding to 0\textsuperscript{th}
step and 20 corresponding to 40\textsuperscript{th} step. This also
places the original unmodified weights on [20,20] point in the
middle of the plot, as can be seen in Figure~\ref{fig:2-1}. This range choice has
allowed us to observe unexpected behavior of NNs with batch
normalization (BN) discussed in Section~\ref{batch-norm-explosion}.

\subsection{Choosing direction vectors for landscape plotting}\label{dir-vects}

At first glance the idea to use random directions in high-dimensional
space might not seem very promising, since one might expect to get very
different results depending on the choice of directions. However, the
landscapes plotted along random directions are not as ``random'' as one
might think. It was previously shown that the number of non-degenerate
directions in parameter space decreases drastically during training~\cite{4H}.
Therefore, there is only a handful of directions that allows
to obtain significantly different results. That being said, finding
multiple directions to obtain different perspectives on the landscape
might indeed be challenging, which is one of the shortcomings of random
axes method~\cite{Hd}.

Fortunately, there are several deterministic methods to choose the
directions. The first one has to do with the underlying geometry of the
parameter space, namely the Hessian of the neural network weight matrix~\cite{Hd}, 
which will be discussed in detail in Section~\ref{hes-ax-esd}. The
information about the landscape one obtains with Hessian axes is much
richer, as will be shown in the next subsection. Another way is to use
optimizer parameter axes as vectors. For models trained with Adam
optimizer~\cite{Adm}, these parameter vectors are moment vectors that
influence the way model weights are altered during training. This allows
to obtain significantly different loss landscapes analyzing which is out
of the scope of this paper. The options to use Hessian and Adam axes for
loss landscape plotting is implemented in LLA library.

\subsection{Normalization of direction vectors}\label{normalization}

Normalization of direction vectors is often necessary since random
values between 0 and 1 are added to weights of undefined scale in~\ref{eq:1}.
It is also desired that contributions of different layers and filters to
loss landscape are comparable, since otherwise some NN elements can
overshadow all others. This would also allow to compare loss landscapes
of NNs with different architectures. Another illustration of the
necessity of normalization is related to the scale invariance of
rectified NNs with ReLU activations~\cite{LLO, 4H}. It has to do with the
observation that if weights of one layer are scaled up (multiplied) by
some amount and weights of the next layer are scaled down (divided) by
the same amount, the output of the neural network will not change.
However, architectures with and without scaled weights will produce
different loss landscapes, making our visualization somewhat arbitrary.

Numerically, normalization is concerned with adjusting the values in
direction vectors depending on weights of the studied neural network.
There are four normalization methods which include weight~\cite{LLOg},
filter~\cite{LLO}, layer~\cite{LLO}, and model~\cite{LLg} normalization
methods, all of which are realized in LLA. The first one is the most
intuitive yet fails to account for scale invariance. The latter two
normalize the direction of layer and model vectors, but do not consider
individual filters. Filter normalization proposed in~\cite{LLO} seemingly
satisfies all requirements of a good normalization method. However, in
this paper we report that filter normalization, along with other method,
does not prevent ``value explosion'' observed for some NNs. This is
because while normalization prevents the uncontrollable growth of
weights, it does not prevent layer outputs from changing uncontrollably.
This ultimately means that the necessity to propose an optimal
normalization method still exists.

\subsection{Batch norm layers and ``value explosion'' in loss landscapes}\label{batch-norm-explosion}

Considering the discussion above, one might expect BN to be a natural
solution since it normalizes the outputs of intermediate layers.
However, one first has to consider that BN layers behave~\cite{BN}
differently in \emph{train} and \emph{eval} operating modes of neural
networks~\cite{BNt}. BN statistics are calculated for every input batch
and used to learn BN parameters in \emph{train} mode, which are later
applied during inference in \emph{eval} mode. Therefore, layer output
normalization is guaranteed only in \emph{train} mode, whereas the
results in \emph{eval} mode depend on the similarity between training
and evaluation data.

Furthermore, when we modify weights to plot loss landscapes, we
potentially make learned BN weights irrelevant. This behavior was
previously observed by other researches who proposed to not modify BN
weights with direction vectors~\cite{LLO}. However, in our experiments
this approach did not yield any positive results, which can partly be
due to LLA plotting loss landscapes for a rather vast region surrounding
the point corresponding to the original weights. The most robust
solution is plotting loss landscapes in \emph{train} mode allowing BN
layers to work as intended preventing value explosions. This is a
problem since \emph{eval} mode is the main inference mode of neural
networks used in all applications ``in the Wild''.

That being said, one could argue that such effects of BN on loss
landscapes is a mere artifact of the method. In real inference scenarios
the changes in dataset features cannot really lead to NN value explosion
due to preprocessing, input data normalization, etc. Hence, value explosion
in loss landscapes has to do only with the way we alter the weights in~(\ref{eq:1}),
and using BN in \emph{train} mode allows NN to adjust to these
changes similar to how preprocessing works for unseen data. Therefore,
using \emph{train} mode gives results that are actually more meaningful
for the analysis of inference modes. However, this argument does not
resolve the ambiguity between \emph{train} and \emph{eval} mode plotting
completely, illustrating the need for a new normalization method.

\subsection{Loss landscapes of popular NNs}\label{ll-pop-nns}

In this Section we examine loss landscapes of popular networks to
investigate the applicability of the existing visualization methods.
These are obtained for randomly initialized and ImageNet-1K pre-trained
NNs. All models and their pre-trained weights are taken from PyTorch
library~\cite{TvM} with the exception of VIT that was trained as part of this study.

\begin{figure} 
    \centering
    \includegraphics[width=\textwidth]{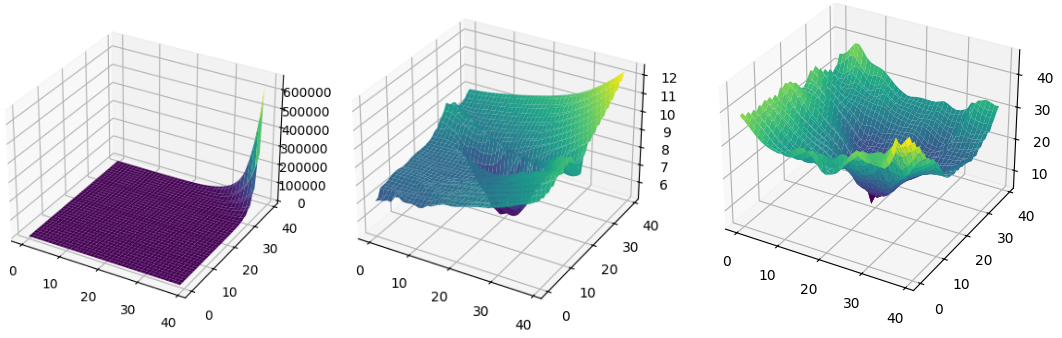}
    \caption{Loss landscapes for ResNet18 in different regimes: (left)
    value explosion in \emph{eval} mode, (center) normal behavior with
    random axes in \emph{train} mode, (right) normal behavior with hessian
    axes in \emph{train} mode.}\label{fig:2.4-1}
\end{figure}

Table~\ref{tab:2-4} shows that value explosion is observed for most CNNs in
various regimes. Figure~\ref{fig:2.4-1} shows that when value explosion occurs, no
meaningful information can be obtained unless a regime which avoids it
is chosen. In addition to \emph{train} mode, changing axes from random
to Hessian sometimes solves this problem. Table~\ref{tab:2-4} also shows that
there is no normalization mode that allows to avoid value explosion
completely. Furthermore, filter normalization always leads to value
explosion if the NN is prone to it. Using \emph{L1} instead of \emph{L2}
norm in filter normalization can sometimes help to avoid this.

\begin{figure} 
    \centering
    \includegraphics[scale=0.5]{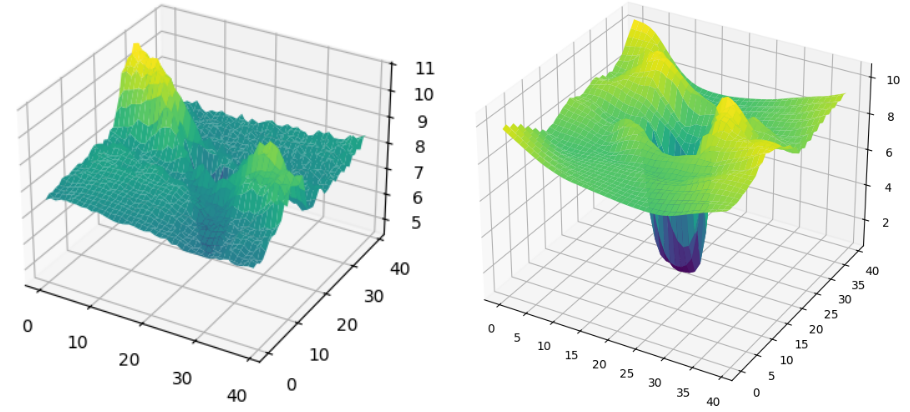}
    \caption{(Left) MobileNet and (right) VIT loss landscapes with
    filter \emph{L2} normalization which exhibit no ``value explosion''.}\label{fig:2.4-2}
\end{figure}

\begin{figure} 
    \centering
    \includegraphics[scale=0.5]{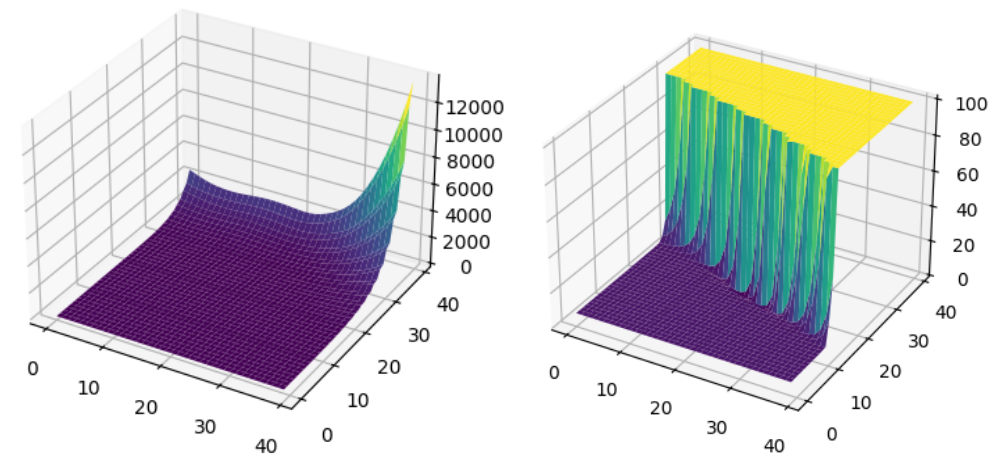}
    \caption{Filter-normalized loss landscapes of models that exhibit
    value explosion, typical for SqueezeNet, AlexNet, and LeNet: (left) no
    loss cap and (right) loss caped at 100.}\label{fig:2.4-3}
\end{figure}

It should be noted that MobileNet and VIT whose landscapes are shown in
Figure~\ref{fig:2.4-2} are the only studied NNs that do not exhibit value
explosion in any mode. Loss landscapes of AlexNet~\cite{AN}, SqueezeNet~\cite{SN}, 
and LeNet~\cite{LN} behave almost identically and have
asymmetric minimum when value explosion occurs, which can be viewed when
loss is capped at some value, as shown in Figure~\ref{fig:2.4-3}.

\begin{table}
    \caption{The behavior of loss values in loss landscapes of CNNs and VIT; digits refer to normalization methods: 
    1 -- none, 2 -- weight -- 3 filter L1, 4 -- filter L2.}\label{tab:2-4}
    \begin{tabular}{|l|l|l|l|l|}
      \hline
      model & Explosion in \emph{train},  & Explosion in
      \emph{train},  & Explosion in \emph{eval}, &
      Explosion in \emph{eval}, \\
      & random axes & Hessian axes & random axes & Hessian axes \\
      \hline
      LeNet & yes\textsuperscript{134} & no\textsuperscript{234} &
      yes\textsuperscript{134} & no\textsuperscript{234} \\
      \hline
      ResNet & no & no & yes & yes\textsuperscript{234} \\
      \hline
      AlexNet & yes\textsuperscript{134} & yes\textsuperscript{24} &
      yes\textsuperscript{134} & yes\textsuperscript{24} \\
      \hline
      SqueezeNet & yes\textsuperscript{124} & yes\textsuperscript{24} &
      yes\textsuperscript{124} & yes\textsuperscript{24} \\
      \hline
      MobileNet & no & no & no & no \\
      \hline
      VIT & no & no & no & no \\
      \hline
    \end{tabular}
    yes\textsuperscript{n...m} means that only normalization methods n...m
    lead to value explosion,

    no\textsuperscript{n...m} means that for normalization methods n...m
    values do increase significantly (but do not explode).
  \end{table}

\section{Hessian of neural networks and Hessian analysis criteria}\label{hes-section}

In context of NNs, Hessian is a matrix of second derivatives of loss
function with respect to weights that incorporates all information about
the curvature of loss function at a point. Most importantly, Hessian
eigenvalues can provide valuable information about local curvature of
loss landscapes. Since forming Hessian explicitly is a very costly task,
the applicability of Hessian analysis and other second order methods is
somewhat limited. However, owing to the recent advances in Randomized
Numeric Linear Algebra (RandNLA)~\cite{R1, R2}, properties of the Hessian
which include its eigenvalues, trace, and eigenvalue spectral density
can be evaluated stochastically even when Hessian matrix is not
available~\cite{L1,PyH}. This facilitates its application to neural
network analysis problems.

\subsection{Hessian axes and spectral density}\label{hes-ax-esd}

Conventionally, Hessian eigenvalues and trace have been the most common
criteria for Hessian analysis~\cite{4H, PyH}. It was observed that
eigenvalues and trace grow for poorly trained networks making their
values a possible indicator of network's performance. However, absolute
values of these parameters depend on specifics of NN architectures and
weight values, which, as has been discussed in Section~\ref{hes-ax-esd}, do not
necessarily have to be unique for NNs to perform identically. This makes
the use of eigenvalues and trace somewhat unreliable especially when
different networks are to be compared.

However, there is a very clear geometric relation between Hessian
eigenvalues and loss landscape curvature. That is, negative eigenvalues
are indicative of the presence of areas with negative curvature, whereas
positive ones correspond to the areas of positive curvature. That
inspired some researchers to propose the ratio of negative and positive
eigenvalues to be a measure of local curvature of loss landscapes~\cite{LLO}
which unfortunately requires evaluating Hessian eigenvalues
at every point of the landscape. Another approach is the evaluation of
the full Hessian Eigenvalues Spectral Density (HESD), which provides
information about all Hessian eigenvalues~\cite{PyH}. This is useful since the sole
presence of large eigenvalues may give limited information in case that
the ``weight'' of this eigenvalue (and the direction of associated
eigenvector) is relatively small in the spectrum.

In this paper we mainly focus on HESD evaluation as the main tool for
Hessian analysis. It is based on Stochastic Lanczos Algorithm and for
the exact derivation the reader is referred to~\cite{L1,PyH,PyHg}. LLA
also provides additional useful applications for Hessian eigenvectors,
as they can be used as deterministic direction vectors for weight update
in~(\ref{eq:1}). Specifically, LLA uses two eigenvectors that correspond to two
largest Hessian eigenvalues as Hessian directions. All results which use
``Hessian axes'' in this paper have been obtained using this method.
This method is similar to the one proposed in~\cite{Hd} and allows to
obtain useful results while avoiding some drawbacks of random direction
method, as Section~\ref{methodology} showed.

\subsection{Typical HESD types}\label{typical-esd}

When researching HESD application to NN analysis, one might find similar
HESDs for different NN architectures reported by different research
teams~\cite{4H, PyH, HGPT}. This observation inspired us to conduct an
investigation into HESDs of different networks that concluded that their
structure is almost universal. This Section aims to list typical HESDs
and provide an explanation to this phenomenon.

\begin{figure} 
    \centering
    \includegraphics[width=\textwidth]{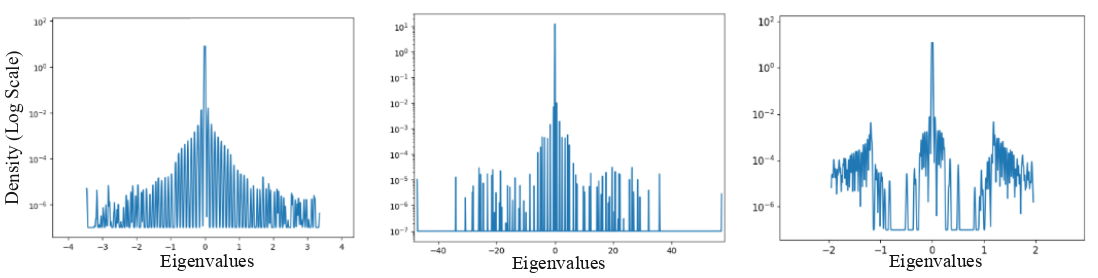}
    \caption{HESD plots of untrained neural networks with randomly initialized weights: (left) AlexNet, 
    (center) SqueezeNet 1.1, and (right) MobileNetV2.}\label{fig:3.2-1}
\end{figure}

The first observation is that HESD plots of untrained NNs with weights
randomly chosen from a uniform distribution are symmetrical with respect
to zero. Figure~\ref{fig:3.2-1} illustrates this for several CNNs, and this
behavior has also been observed for much larger class of NNs. Whereas
the exact shape of HESD is defined by network architecture, the symmetry
is a consequence of symmetrical structure of weight matrices which, when
filled with uniformly distributed random numbers, yields no preferred
direction in Hessian space resulting in a symmetric spectrum.

\begin{figure} 
    \centering
    \includegraphics[scale=0.5]{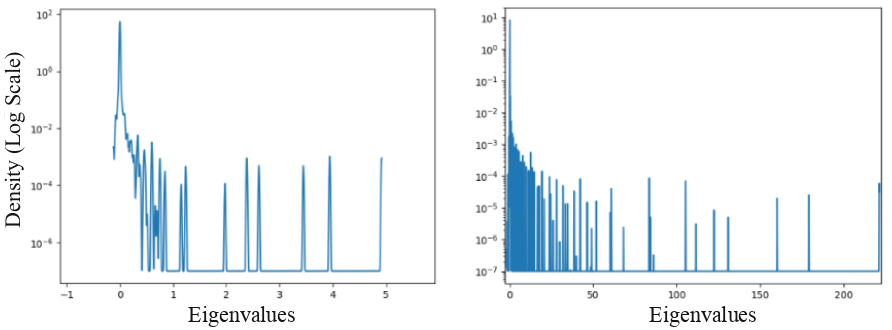}
    \caption{HESD plots of neural networks trained to over 99\% accuracy: (left) 
    LeNet trained on MNIST and (right) ResNet20 trained on Cinic10.}\label{fig:3.2-2}
\end{figure}

The second observation is that a fully trained neural network with very
low training loss and high training accuracy has almost exclusively
positive eigenvalues, as shown in Figure~\ref{fig:3.2-2}. This can be explained by
the fact that during training optimizers aim to change model weights
along the negative slope of the loss landscape. Hence, for a fully
trained NN there is essentially no negative slope left, which is
reflected in the near absence of negative eigenvalues in HESD.

\begin{figure} 
    \centering
    \includegraphics[width=\textwidth]{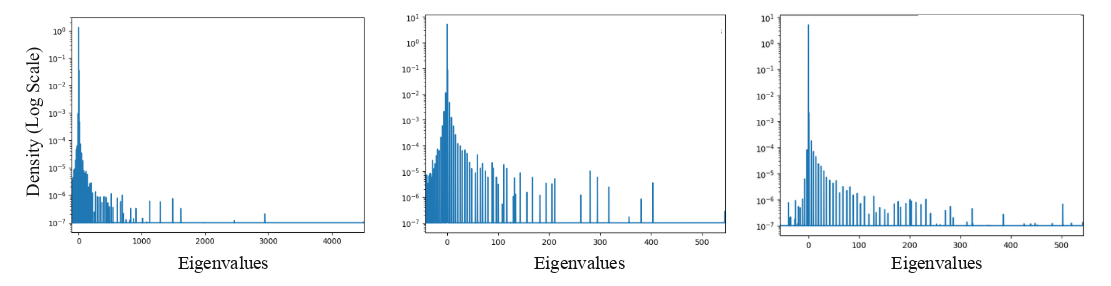}
    \caption{HESD plots of neural networks trained to 70-80\% 
    accuracy: (left)  SqueezeNet 1.1, (center) MobileNetV2, 
    and (right) VIT trained on ImageNet-1K.}\label{fig:3.2-3}
\end{figure}

Finally, in intermediate states between random initialization and being
fully trained, i.e., during training, the negative eigenvalue section of
HESD reduces while the positive section grows. This is illustrated by
Figure~\ref{fig:3.2-3} and other results obtained in this study. This behavior can
also be explained by the gradual reduction in negative eigenvalues
during training accompanied by gradual emergence of large positive
eigenvalues. The latter were previously referred to as outliers which
correspond to high positive curvature directions associated with
classifier classes of studied NNs~\cite{4H}. Near-zero eigenvalue
sections also correspond to a large number of degenerate directions in
Hessian space which indicates a mostly flat loss landscape.

Such HESD plots have been observed for a large class of convolutional
and fully-connected NNs. Similar HESD plots were also observed for VITs and
Generative Pretrained Transformers (GPTs) \cite{HGPT}. It was suggested
that typical HESD structures can be attributed to block structure of NNs
and the same loss function (cross-entropy) used for all classifiers, which
also holds true for our results.

\begin{figure} 
    \centering
    \includegraphics[width=\textwidth]{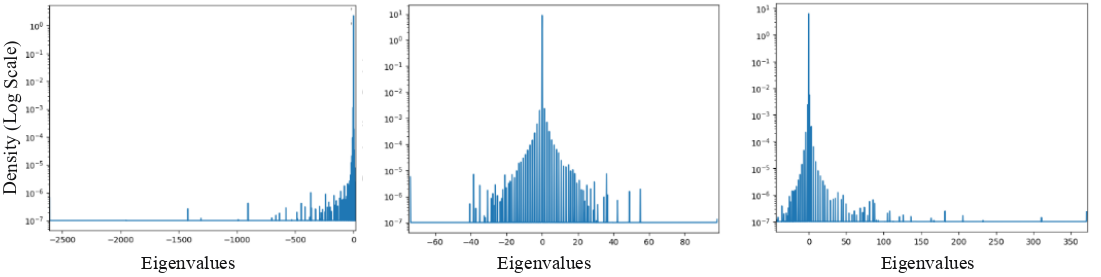}
    \caption{HESD plots of randomly initialized ResNet18 in (left)
    \emph{train} and (center) \emph{eval} modes, and (right) its ImageNet-1K
    pre-trained version.}\label{fig:3.2-4}
\end{figure}

However, two exceptions to the above behavior have been observed. First,
ResNets have asymmetric HESD plot upon random initialization, as shown
in Figure~\ref{fig:3.2-4}. This effect occurs only in \emph{train} mode of the
neural network when BN parameters are calculated using input batch data
statistics, and it is not present for evaluation mode when all
parameters are fixed. This means that the asymmetry which arises from
parameter calculation in BN
layers is sufficient to create relatively large negative eigenvalue
sections in HESD. However, this effect is either minor or not present in
other NNs with BN layers.

\begin{figure} 
    \centering
    \includegraphics[scale=0.5]{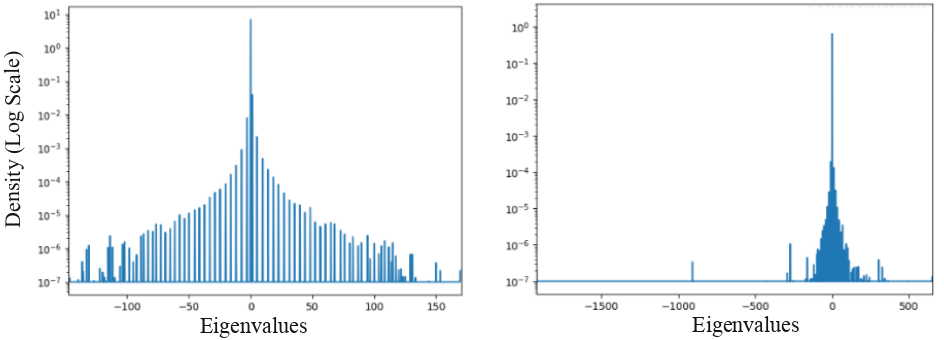}
    \caption{HESD plot of (left) randomly initialized and (right) pre-trained VIT from~\cite{VTg,VTtrain}.}\label{fig:3.2-5}
\end{figure}

Lastly, HESD of pre-trained VIT~\cite{VTg} shown in Figure~\ref{fig:3.2-5} is different from the
previously discussed ones. It is characterized by a large negative
eigenvalue section, even that this VIT has a relatively high training
accuracy and low loss. It is likely to be related to the differences in training of this specific VIT~\cite{VTtrain} and
other models discussed in this study. It requires additional
study and the detailed analysis of this VIT's behavior will be conducted later.

\subsection{Hessian analysis criteria}\label{hes-crit}

In this Section we propose several criteria that can be
conveniently calculated using HESD to analyze it without necessarily
having to visually inspect the plots. Following the discussion in
Section~\ref{hes-ax-esd}, the first criterion is simply the ratio of most negative
and most positive Hessian eigenvalues

\begin{equation}
    r_{\text{e}} = \frac{\max\left(\mathrm{abs}\left( \lambda_{\text{neg}} \right)\right)}{\max\left( \lambda_{\text{pos}} \right)},
    \label{eq:2}
\end{equation}

where \(\lambda_{\text{neg}}\) and \(\lambda_{\text{pos}}\) are negative and
positive Hessian eigenvalues, respectively. However, this
criterion does not take into account the complete eigenvalue spectrum.
To address this issue, the following metric is proposed

\begin{equation}
    K_{\text{Hn}} = \frac{\sum\limits_{i}^{}\left(\mathrm{abs}\left( \lambda_{\text{neg,i}}\right) \cdot w_{\text{neg,i}} \right)^n}
    {\sum\limits_{j}^{}\left( \lambda_{\text{pos,j}} \cdot w_{\text{pos,j}} \right)^n},
    \label{eq:3}
\end{equation}

where \(\omega_{\text{neg}}\) and \(\omega_{\text{pos}}\) are weights
corresponding to negative and positive eigenvalues used to form HESD
(see Section 3C in~\cite{PyH} for details), respectively, and \emph{n} is
some positive real number.

To investigate whether these criteria reflect the changes in HESD and NN
performance correctly, we calculate \emph{r\textsubscript{e}},
\emph{K\textsubscript{H1}} (\emph{n} = 1 in~(\ref{eq:3})), and
\emph{K\textsubscript{H05}} (\emph{n} = 0.5 in~(\ref{eq:3})) for several neural
networks and compare their random initialization states to the
pre-trained ones, as shown in Table~\ref{tab:3.3-1}. ImageNet-1K pre-trained
weights were used for all neural networks but LeNet, which was trained
on MNIST as part of this study. Accuracy is calculated for the same
batch of training data that was used to evaluate HESD of pre-trained
models and averaged for \emph{train} and \emph{eval} experiments.

\begin{table}
    \caption{HESD criteria for randomly initialized and pre-trained
    neural networks. Three values in cells correspond to
    \emph{K\textsubscript{H1}} , \emph{K\textsubscript{H05}} , and
    \emph{r\textsubscript{e}}, respectively.}\label{tab:3.3-1}
    \begin{tabular}{|l|l|l|l|l|l|}
      \hline
      model & Random init, & Pre-trained, & Random init, & Pre-trained, & Average \\
            & \textit{train} mode & \textit{train} mode & \textit{eval} mode & \textit{eval} mode & accuracy \\
        \hline
            
        Lenet & 0.99, 0.95, 0.88 & 0.37, 0.23, 5e-4 & 0.98, 0.98, 0.8 & 0.18,
        0.17, 3e-4 & 99 \%\\ \hline
        ResNet18 & 1.15, 2.3, 451 & 1.03, 0.67, 0.32 & 0.97, 0.87, 0.73 & 0.97,
        0.66, 0.03 & 86 \%\\ \hline
        AlexNet & 1.0, 1.0, 1.0 & 0.94, 0.64, 0.04 & 0.99, 0.97, 1.0 & 0.97,
        0.67, 0.06 & 84 \%\\ \hline
        SqueezeNet & 1.18, 1.23, 0.96 & 0.7, 0.43, 0.02 & 1.08, 1.23, 1.01 &
        0.76, 0.55, 0.02 & 75 \%\\ \hline
        MobileNet & 1.07, 1.21, 5.7 & 1, 0.75, 0.1 & 1.04, 1.0, 1.0 & 0.83,
        0.54, 0.01 & 84 \%\\ \hline
        VIT & 0.99, 0.96, 0.87 & 0.86, 0.47, 0.04 & 0.99, 0.99, 0.87 & 0.85,
        0.46, 0.04 & 98 \%\\ \hline
    \end{tabular}
  \end{table}

Table~\ref{tab:3.3-1} shows that whereas \emph{K\textsubscript{H1}} works for
LeNet trained to very high accuracy, it fails to capture the changes
between pre-trained and randomly initialized states of other CNNs. On
the contrary, \emph{K\textsubscript{H05}} consistently decreases when
pre-trained weights are used. That being said, there is no direct
correlation between \emph{K\textsubscript{H05}} and accuracy values,
since for networks with 85\% average accuracy
\emph{K\textsubscript{H05}} ranges from 0.64 to 0.75. Therefore, it is
the change in \emph{K\textsubscript{H05}} value but not its exact value
that can be used to access neural network's performance. It is also
important that it allows to obtain some estimate of NN performance while
requiring only a few (even just one) batches of input data for its
calculation.

Table~\ref{tab:3.3-1} also shows that \emph{r\textsubscript{e}} decreases for
pre-trained models, too. However, its variation range is extremely large
making it less convenient to use as a criterion compared to
\emph{K\textsubscript{H}}. It also sometimes does not reflect the
changes in accuracy correctly, as will be shown in Section~\ref{gen-exp}. On the
other hand, it can help to capture abnormalities like the behavior of
randomly initialized ResNet18 in \emph{train} mode, for which
\emph{r\textsubscript{e}} sky-rockets to 451. 

\subsection{Computational speed and stability of HESD criteria}\label{crit-stability}

Since HESD is calculated using stochastic methods, it can vary
depending on calculation seed. Our tests have shown that Lanczos part of
HESD evaluation algorithm is, predictably, the source of results'
variations. Therefore, the criteria proposed in the previous Section can
also vary, which brings their reliability into question.

It was determined that weights in~(\ref{eq:3}) are affected by stochastic nature
of the algorithm more than eigenvalues in~(\ref{eq:2}) and~(\ref{eq:3}), making
\emph{r\textsubscript{e}} a more computationally stable criterion
compared to \emph{K\textsubscript{H}}. Regarding the latter, several
experiments were conducted to investigate the optimal conditions for
\emph{K\textsubscript{H}} application. NVIDIA A100 GPU was used for all
experiments.

First, the dependence of the results on the number of HESD evaluation
runs (\emph{n\_hes}) was studied. When running consequent calculations
with different seeds, for instance, for average
\emph{K\textsubscript{H05}} = 0.16 with variation is 0.13-0.2.
\emph{n\_hes} = 10 allows to get relatively stable results which vary
slightly with later calculations. Increasing \emph{n\_hes} to 20 allows
to obtain more reliable results, though it is more computationally
expensive.

Another aspect that is not connected with stochastic uncertainties but
can affect the results is the choice of data for HESD calculation. While
one batch of data is sufficient to calculate the criteria, the results
may vary for different batches in the dataset. It was noticed that even
for high \emph{n\_hes} the variation can be substantial, for instance,
for average \emph{K\textsubscript{H05}} = 0.16 the variation is 0.1-0.24
for LeNet trained on MNIST. However, averaging for \emph{N} random
batches allows to get relatively stable results. For the same example
with \emph{K\textsubscript{H05}} = 0.16, the variation drops to 0.05
when at least \emph{N} = 4 batches are used. Similar results were
observed for other studied cases, including ResNet(ImageNet)
experiments. Using several batches also allows to use fewer
\emph{n\_hes}.

However, the necessity to conduct multiple calculations to stabilize the
results questions the efficiency of the proposed method in comparison to
simple accuracy evaluation. For instance, for LeNet on MNIST the
criterion calculation on four batches of 2048 images with \emph{n\_hes}
= 10 takes ten times more time than accuracy calculation on full
dataset. This shows that the proposed criteria are suboptimal for small
dataset and compact networks. On the contrary, for large models and
extremely large datasets like ResNets(ImageNet-1K) the advantages of the
proposed criteria become apparent, since \emph{K\textsubscript{H}}
calculation takes roughly 22 sec for four 64 image batches, which is
equivalent to accuracy calculation on 3500 out of several million
ImageNet images~\cite{IN}.

\section{Generalization experiments}\label{gen-exp}

In this Section we investigate how loss landscape and Hessian analyses
can be used to assess the generalization capabilities of NNs. In order
to achieve this, we use two pairs of datasets with compatible classes to
conduct generalization experiments. This allows NNs trained on one
dataset be used for inference on another one without retraining or
fine-tuning. This setup is more similar to real generalization scenarios
than the setups that use test subset of training dataset to calculate
generalization accuracy~\cite{LLO}.

\subsection{Datasets}\label{datasets}

Two pairs of datasets were used to conduct generalization experiments:
digits' datasets MNIST~\cite{MN} and ``The Street View House Numbers'' (SVHN)~\cite{Hn},
and image datasets Cifar10~\cite{CIF} and Cinic10~\cite{CIN}. All datasets
have 10 classes with 32x32 images, which are grayscale for MNIST and RGB
for others. SVHN images were converted to grayscale to be compatible
with MNIST.

It should be noted that Cinic10 consists of Cifar10 and some ImageNet
images relabeled with Cifar10 class labels. In order to make
generalization experiments valid, all Cifar10 images have been removed
from Cinic10 with this version being referred to as Cinic10-i in Table~\ref{tab:ge-1}. 
Validation split of Cinic10 has not been included into Cinic10-i.
For simplicity we will refer to Cifar10 as cifar and to Cinic10-i as
cinic in this paper. All datasets and their train/test splits are
described in Table~\ref{tab:ge-1}.

\begin{table}
    \caption{Description of the datasets used in this study.}\label{tab:ge-1}
    \centering
    \begin{tabular}{|l|l|l|l|l|l|}
      \hline
      Dataset & Image shape & Dataset size & Train split & Test split & Note \\ 
      & & & size & size & \\ \hline

      MNIST & 32x32x1 & 70000 & 60000 & 10000 & -\\ \hline
      SVHN & 32x32x1 & 99289 & 73257 & 26032 & Converted to grayscale \\ 
      & & & & & from original RGB.  \\ \hline
      Cifar10 & 32x32x3 & 60000 & 50000 & 10000 & -\\ \hline
      Cinic10 & 32x32x3 & 270000 & 90000 & 90000 & Cifar10 and ImageNet \\ 
      & & & & & images  \\ \hline
      Cinic10-i & 32x32x3 & 140000 & 70000 & 70000 & ImageNet images
      only\\ \hline
    \end{tabular}
  \end{table}

In addition, one batch of 64 images was collected from the internet and manually
labeled with ImageNet-1K labeles to test generalization capabilities of
popular models pre-trained on ImageNet. To minimize the odds
that some of the images were part of ImageNet training subset, only
relatively recent images not used on popular websites were considered.
Images were rescaled to 224x224x3 to match the input size requirements
of the pre-trained models. This dataset is referred to as NI in Section~\ref{popnets-gen}.

In all following experiments models are trained for 1000 epochs with
0.001 learning rate using Adam optimizer on the train split of the
training dataset which is used for training accuracy calculation. When
reporting results, training dataset is indicated in brackets after NN
name, e.g. MNIST in LeNet(MNIST). Generalization experiments consist of
conducting model inference on the test subset of another dataset, and
then calculating generalization accuracy on that subset. This dataset is
indicated after dash symbol, e.g. SVHN in LeNet(MNIST)-SVHN. The
experiments constituting of NN training and testing on the same dataset
are not included to avoid confusion.

\subsection{LeNet MNIST-SVHN generalization experiments}\label{lenet-gen}

\begin{figure} 
    \centering
    \includegraphics[width=\textwidth]{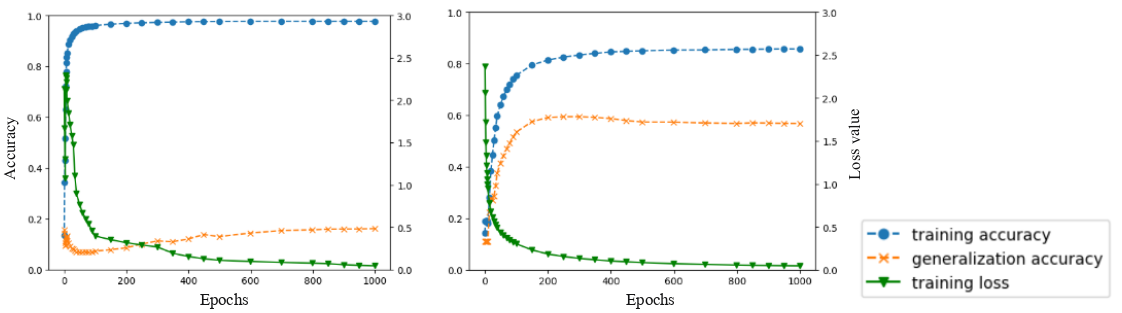}
    \caption{Training loss, training accuracy, and generalization accuracy (left) 
    LeNet(MNIST)-SVHN and (right) LeNet(SVHN)-MNIST.}\label{fig:4.2-1}
\end{figure}

The first dataset pair was used to train LeNet. Figure~\ref{fig:4.2-1} shows that
generalization accuracy is much lower when generalizing from MNIST to
SVHN. This behavior can be expected since a smaller dataset is used for
training. LeNet(MNIST)-SVHN also shows abnormal growth in
generalization accuracy after the training is essentially finished,
which is not observed in other experiments. However, this increase
amounts only to several percent not even reaching 20\% generalization
accuracy at 1000\textsuperscript{th} epoch.

On the contrary, LeNet exhibits 55-58\% generalization accuracy on MNIST
when trained on SVHN. It should be noted that higher generalization
accuracy is accompanied by lower training accuracy that does not reach
90\%. In both cases loss continues to decrease even after maximum
accuracy is reached, and for LeNet(SVHN) experiment this does not have
any significant effect on generalization accuracy.

As has been previously shown in Table~\ref{tab:2-4}, LeNet loss landscape exhibits
value explosion in many regimes which limits the applicability of loss
landscape analysis to its generalization capability evaluation. Figure~\ref{fig:A1}
shows loss landscapes for LeNet(MNIST) with weight normalization
using random axes that show drastic increase in loss value when tested
on SVHN. However, loss values seem to generally be lower for MNIST,
since for generalization landscape LeNet(SVHN)-MNIST loss values
decrease, too. This makes loss landscape analysis in this case ambiguous
because no direct correlation between landscape changes and
generalization accuracy can be derived.

\begin{figure} 
    \centering
    \includegraphics[scale=0.5]{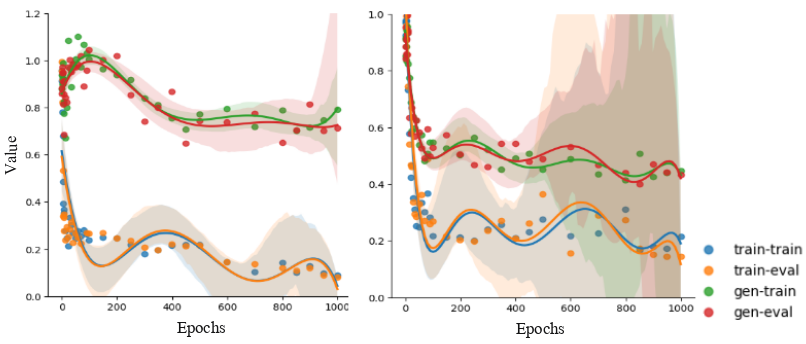}
    \caption{The results for \emph{K\textsubscript{H05}} for (left)
    LeNet(MNIST)-SVHN and (right) LeNet(SVHN)-MNIST.}\label{fig:4.2-2}
\end{figure}

However, Figure~\ref{fig:4.2-2} shows that \emph{K\textsubscript{H05}} increases
significantly when changing datasets. Furthermore, this correlates with
changes in generalization accuracy, since \emph{K\textsubscript{H05}}
increases (and accuracy drops) for MNIST-SVHN case much more
drastically, than for SVHN-MNIST case. Plots for other criteria are
shown in Appendix~\ref{app-B}.

\subsection{ResNet20 cifar-cinic generalization experiments}\label{resnet-gen}

\begin{figure} 
    \centering
    \includegraphics[width=\textwidth]{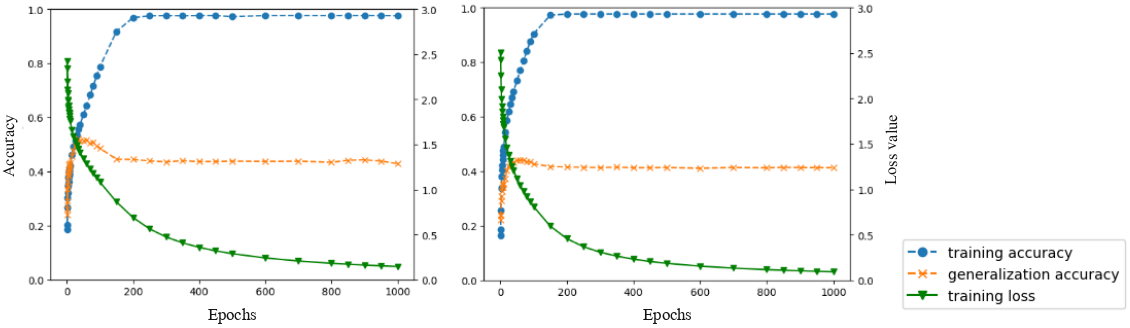}
    \caption{Training loss, training accuracy, and generalization
    accuracy of (left) ResNet20(cinic)-cifar and (right)
    ResNet20(cifar)-cinic.}\label{fig:4.3-1}
\end{figure}

Figure \ref{fig:4.3-1} shows that the final training accuracy of approximately 97\% is
reached around 200\textsuperscript{th} epoch for both ResNet20(cinic)
and ResNet20(cifar). However, generalization accuracy reached its peak
value at approximately 70\textsuperscript{th} epoch and then decreased
to around 43\% maintaining this value with roughly 1\% oscillation for
both experiments. The highest generalization accuracy of 55\% is reached
by ResNet(cinic)-cifar, while ResNet(cifar)-cinic genealization accuracy
does not exceed 45\%.

It is interesting that unlike MNIST-SVHN experiment where training on
smaller dataset results in much worse generalization accuracy,
ResNet20(cifar)-cinic reaches roughly the same training and
generalization accuracies as ResNet20(cinic)-cifar. This might be
explained by very similar feature distributions in both datasets~\cite{CIN}.

\begin{figure} 
    \centering
    \includegraphics[scale=0.5]{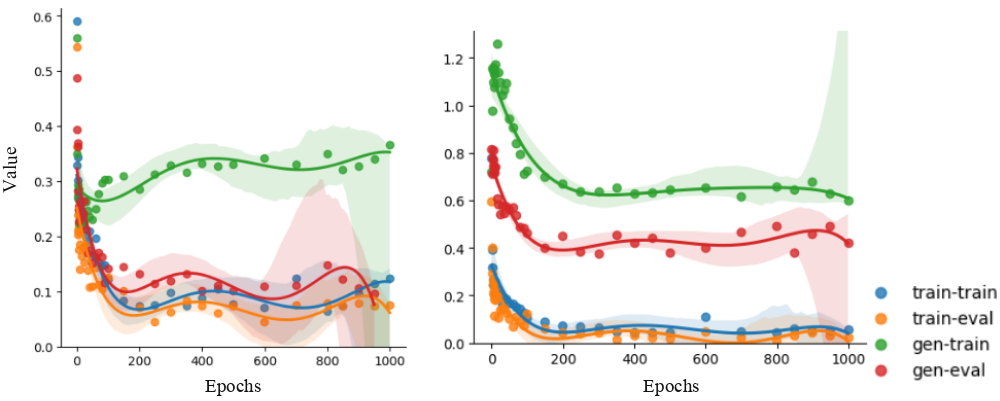}
    \caption{The results for criterion \emph{K\textsubscript{H05}} for
    (left) ResNet20(cinic)-cifar and (right) ResNet20(cifar)-cinic.}\label{fig:4.3-2}
\end{figure}

Figure~\ref{fig:A2} shows that loss landscapes of ResNet20 are overall rather
non-uniform even when plotted for training data. Therefore, even that
the cifar-cinic plot is clearly more chaotic than the cifar training
one, it is hard to give quantitative estimates to the changes in NN
performance. On the contrary, the results for criterion
\emph{K\textsubscript{H05}} shown in Figure~\ref{fig:4.3-2} indicate that it
increases in generalization experiments, thus corresponding to the
decrease in accuracy. However, the \emph{r\textsubscript{e}} criterion
is smaller for generalization than \emph{r\textsubscript{e}} for
training, not reflecting the accuracy changes correctly, as figures in
Appendix~\ref{app-B} show.

This means that the increase in \emph{K\textsubscript{H05}} when
changing datasets for a trained model can indicate poor generalization
and the reduction in accuracy. However, as mentioned before, there is no
strict correlation between \emph{K\textsubscript{H05}} and accuracy
values. Indeed, the average \emph{K\textsubscript{H05}} value for
ResNet20(cinic)-cifar is 0.25, and its value for ResNet20(cifar)-cinic
is 0.45, whereas both correspond to approximately 43\% generalization
accuracy.

\subsection{Generalization capability of popular NNs}\label{popnets-gen}

Table~\ref{tab:4.4-1} summarizes the results of generalization experiments
conducted with several popular CNNs and VIT. One random batch of 64
ImageNet-1K images is used to calculate ImageNet HESD criteria and
accuracy. In all cases ImageNet-1K pre-trained weights were used for
model inference.

\begin{table}
    \caption{HESD criteria and inference accuracy of popular CNNs and VIT.}\label{tab:4.4-1}
    \centering
    \begin{tabular}{|l|l|l|l|l|l|}
      \hline
      model & dataset & Accuracy, \% & \emph{K\textsubscript{H1}} &
      \emph{K\textsubscript{H05}} & \emph{r\textsubscript{e}} \\ \hline
      ResNet50 & ImageNet & 85.9 & 0.86 & 0.48 & 0.064 \\ \hline
      ResNet50 & NI & 71.9 & 0.95 & 0.65 & 0.065 \\ \hline
      AlexNet & ImageNet & 84.3 & 0.94 & 0.65 & 0.11 \\ \hline
      AlexNet & NI & 23.4 & 0.96 & 0.8 & 0.21 \\ \hline
      SqueezeNet & ImageNet & 75 & 0.72 & 0.5 & 0.06 \\ \hline
      SqueezeNet & NI & 23.4 & 0.88 & 0.64 & 0.086 \\ \hline
      MobileNet & ImageNet & 84.3 & 0.71 & 0.40 & 0.015 \\ \hline
      MobileNet & NI & 59.3 & 0.76 & 0.54 & 0.04 \\ \hline
      VIT & ImageNet & 98.5 & 0.86 & 0.47 & 0.04 \\ \hline
      VIT & NI & 4.9 & 0.99 & 0.93 & 0.49 \\ \hline
    \end{tabular}
  \end{table}

Table~\ref{tab:4.4-1} shows that \emph{K\textsubscript{H05}} behaves as has been
previously discussed in this study, increasing when generalization
accuracy drops for CNNs. As mentioned earlier,
\emph{K\textsubscript{H1}} does not always have this property, as
AlexNet results show. In many cases \emph{r\textsubscript{e}} also
increases when accuracy decreases, though these changes vary
significantly for different experiments, and \emph{r\textsubscript{e}}
behavior does not always reflect that of the accuracy. This makes
\emph{K\textsubscript{H05}} the most suitable Hessian criterion for NN
generalization capability assessment. It should be noted that the studied VIT 
has generalized poorly, which is reflected in hessian criteria. The reasons for
this will be investigated in detail in the future.

\section{Conclusions}\label{conclusions}

This paper discusses the applicability of loss landscape and Hessian
analyses to the study of generalization capabilities of NNs while
focusing on different aspects of the methodology aiming to propose new
analysis criteria. The results were obtained using a Loss Landscape
Analysis library developed as part of this study. Different approaches
to loss landscape evaluation were discussed showing that conventional
methods might fail in specific circumstances which raises the demand for
an improved normalization method. This conclusion was made after a
comprehensive analysis of loss landscapes of conventional NNs in various
regimes was conducted. It was shown that using Hessian axes can
sometimes mitigate this issue, and the approach for choosing Hessian
axes was proposed. Typical Hessian spectra were also discussed and
several criteria for their analysis were proposed. Generalization experiments
conducted using LeNet and ResNet20 on MNIST-SVHN and cifar-cinic
dataset pairs showed that the proposed criteria can be used as an
estimate of generalization accuracy, which is computationally efficient
for large datasets.

\section*{Acknowledgement}\label{acknowledgement}

The author would like to thank his colleagues Dr Anton Raskovalov, Dr
Igor Netay, and Ilya Androsov for fruitful discussions, and Vasily
Dolmatov for discussions and project supervision.

\bibliographystyle{IEEEtran}
\bibliography{IEEEabrv,ms}

\newpage
\appendix
\renewcommand\thefigure{\thesection.\arabic{figure}}    
\section{Appendix A}
\label{app-A}
\setcounter{figure}{0}  

\begin{figure}[ht]
	\centering
	\includegraphics[scale=0.65]{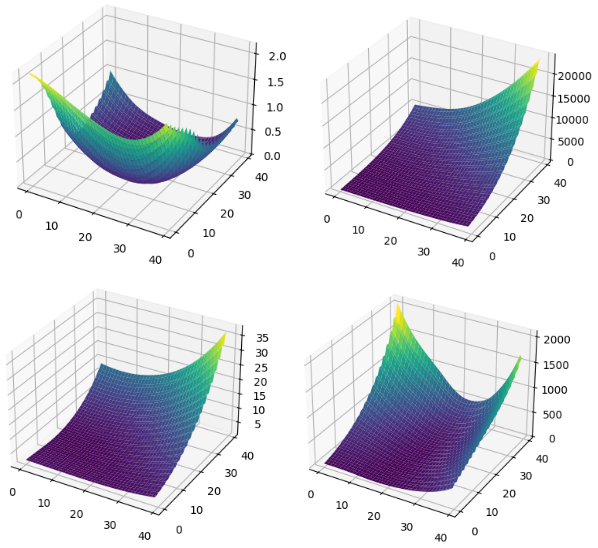} 
	\caption{Random axes weight-normalized loss landscapes of LeNet
    (top-left) trained and tested on MNIST, (top-right) trained on MNIST and
    tested on SVHN, (bottom-left) trained and tested on SVHN, and
    (bottom-right) trained on SVHN and tested on MNIST. All results
    correspond to 249\textsuperscript{th} epoch of respective experiments.}
	\label{fig:A1}
\end{figure}

\begin{figure}
	\centering
	\includegraphics[scale=0.65]{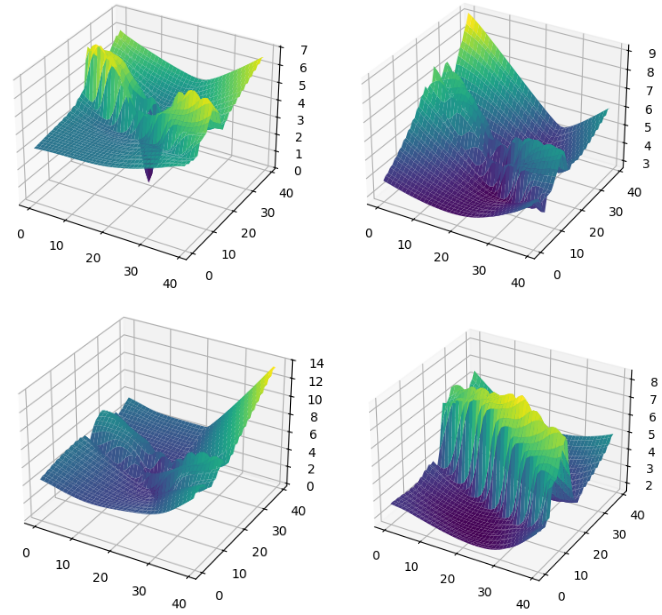} 
	\caption{Random axes filter \emph{L2}-normalized loss landscapes of
    ResNet20 (top-left) trained and tested on cinic, (top-right) trained on
    cinic and tested on cifar, (bottom-left) trained and tested on cifar,
    and (bottom-right) trained on cifar and tested on cinic. All results
    correspond to 249\textsuperscript{th} epoch of respective experiments.}
	\label{fig:A2}
\end{figure}

\newpage
\section{Appendix B}
\label{app-B}
\setcounter{figure}{0}  

\begin{figure}[h]
	\centering
	\includegraphics[width=\textwidth]{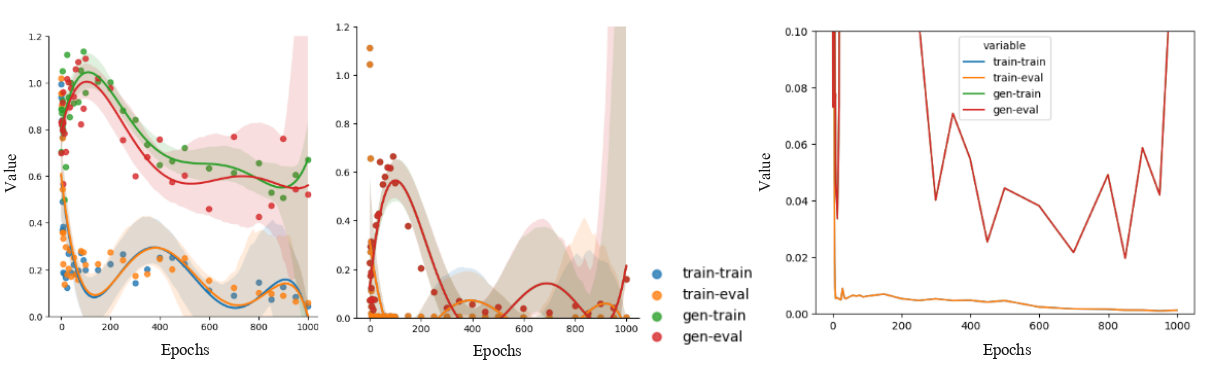} 
	\caption{Hessian criteria for LeNet(MNIST)-SVHN: (left)
    \emph{K\textsubscript{H1}}, (center) \emph{r\textsubscript{e}} from max
    to min values, and (right) \emph{r\textsubscript{e}} in near-zero area.}
	\label{fig:B1}
\end{figure}

\begin{figure}[h]
	\centering
	\includegraphics[width=\textwidth]{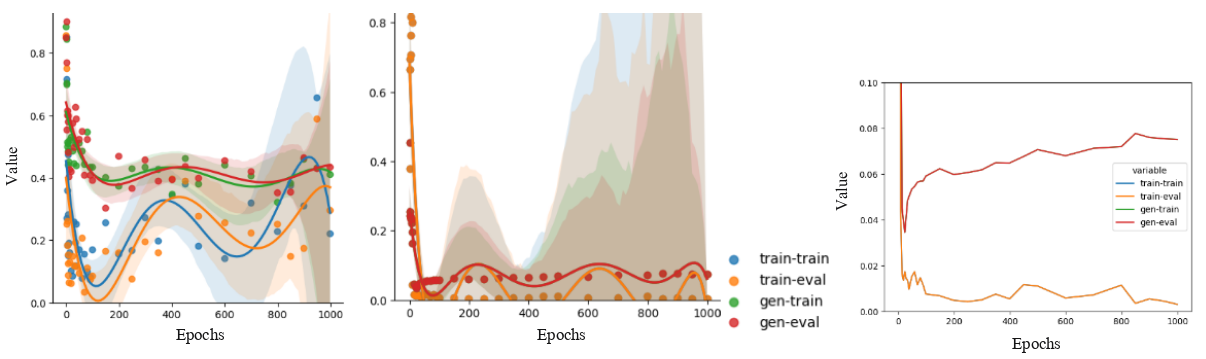} 
	\caption{Hessian criteria for LeNet(SVHN)-MNIST: (left)
    \emph{K\textsubscript{H1}}, (center) \emph{r\textsubscript{e}} from max
    to min values, and (right) \emph{r\textsubscript{e}} in near-zero area.}
	\label{fig:B2}
\end{figure}

\begin{figure}[h]
	\centering
	\includegraphics[width=\textwidth]{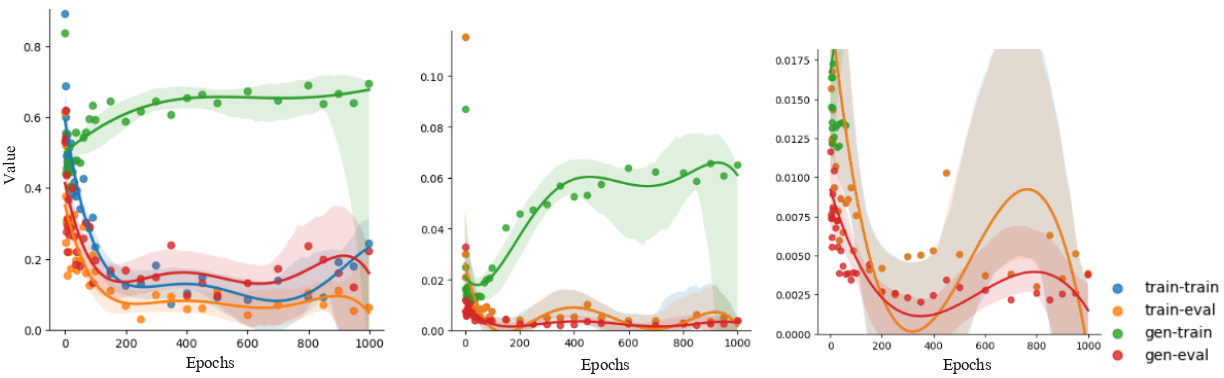} 
	\caption{Hessian criteria for ResNet20(cinic)-cifar: (left)
    \emph{K\textsubscript{H1}}, (center) \emph{r\textsubscript{e}} from max
    to min values, and (right) \emph{r\textsubscript{e}} in near-zero area.}
	\label{fig:B3}
\end{figure}

\begin{figure}[h]
	\centering
	\includegraphics[width=\textwidth]{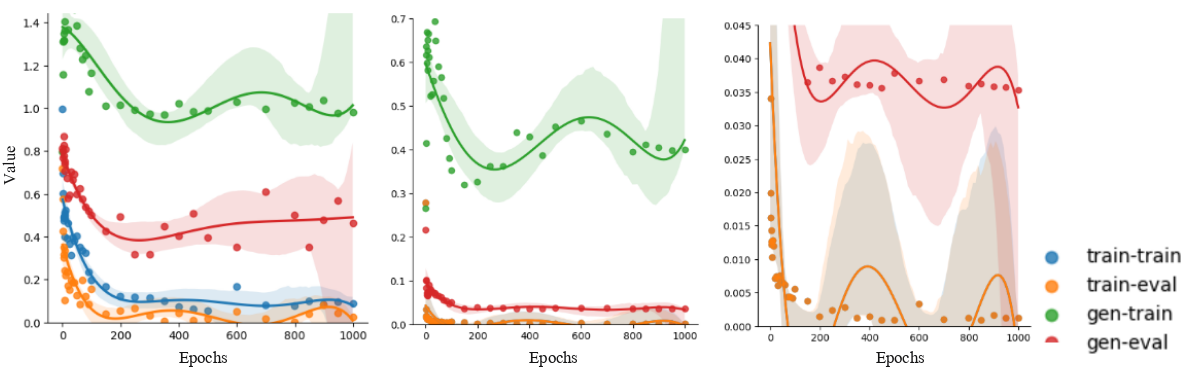} 
	\caption{Hessian criteria for ResNet20(cifar)-cinic: (left)
    \emph{K\textsubscript{H1}}, (center) \emph{r\textsubscript{e}} from max
    to min values, and (right) \emph{r\textsubscript{e}} in near-zero area.}
	\label{fig:B4}
\end{figure}